# Minutiae Based Thermal Human Face Recognition using Label Connected Component Algorithm

Ayan Seal*, Suranjan Ganguly, Debotosh Bhattacharjee, Mita Nasipuri,
Dipak Kumar Basu

*DST INSPIRE FELLOW

Department of Computer Science and Engineering, Jadavpur University, Kolkata- 700032, India

**Abstract**

In this paper, a thermal infra red face recognition system for human identification and verification using blood perfusion data and back propagation feed forward neural network is proposed. The system consists of three steps. At the very first step face region is cropped from the colour 24-bit input images. Secondly face features are extracted from the croped region, which will be taken as the input of the back propagation feed forward neural network in the third step and classification and recognition is carried out. The proposed approaches are tested on a number of human thermal infra red face images created at our own laboratory. Experimental results reveal the higher degree performance.

## 1. Introduction

In the modern society there is an increasingly need to verify and recognize persons automatically. The current technologies of using a password or personal identification number (PIN) for these purposes are not enough since they are moveable, disclosable and hard to memorize. Biometric-based verification and recognition methods are the most dependable. Technological revolution has taken place all over the world based on computer. Computer vision is a part of everyday life. One of the most important goals of computer vision is to achieve visual recognition ability comparable to that of human. By measuring something unique about a human being and using that to identify them. Then we can say that we achieve a vivid improvement in security aspect. Biometrics uses physical characteristic or personal trait to match with the data in the database to determine the possible candidates. Physical feature is suitable for identity purpose and generally obtained from living human body. Commonly used physical features are fingerprints, facial features, hand geometry, and eye features (iris and retina) etc. The most common used personal traits are signature and voices etc; much work is still needed to improve the biometric security systems. Nearly all-biometric systems work in the same manner. First, a person is registered into a database using the specified method. Information about a certain characteristic of the human is captured. When the person needs to be identified, the system will take the information about the person again, translates this new information with the algorithm, and then compare the new code with the ones in the database to find out a match and hence, identification. Among many recognition systems, face recognition has drawn significant attention and interest from many researchers for the last three decades because of its potential applications, such as in the areas of surveillance, Closed Circuit Television (CCTV) control, user authentication, HCI Human Computer Interface, intelligent robot, daily attendance register, airport authority for security checks and immigration checks, police may use for identification of criminals and so on and so forth. Most of the research hard works in this area have focused on visible spectrum imaging due to easy availability of low cost visible band optical cameras. But, it requires an external source of illumination. Performance of visual face recognition is sensitive to variations in illumination conditions and usually degrades significantly when the lighting is not bright or when it is not uniformly illuminating the face. Various algorithms (e.g. histogram equalization, eigenfaces etc.) for compensating such variations have been studied with partial success. To overcome this limitation, several solutions have been designed. One solution is using 3D data obtained from 3D vision device. Such systems are less dependent on illumination changes, but they have some disadvantages: the cost of such system is high and their processing speed is low. Thermal IR images [1] have been suggested as a possible alternative in handling situations where there is no control over illumination. Recently researchers have been using Near-IR imaging cameras for face recognition with better results [2]. Previously Thermal IR camera was costly but recently the cost of IR cameras has been considerably reduced with the development of CCD technology [3]; thermal images can be captured under different lighting conditions, even under completely dark environment. Using thermal images, the tasks of face detection, localization, and segmentation are comparatively easier and more reliable than those in visible band images [4]. The vein and tissue structure of the face is unique for each human being [5] [8], and therefore the IR images are also unique. An infrared camera with good sensitivity can indirectly capture images of superficial blood vessels on the human face [6]. However, it has been indicated by Guyton and Hall [7] that the average diameter of blood vessels is around 10-15 µm, which is too small to be detected by current IR cameras because of the limitation in spatial resolution. The skin just above a blood vessel is on an average 0.1 ◦C warmer than the adjacent skin, which is beyond the thermal accuracy of current IR cameras. The convective heat transfer effect from the flow of "hot" arterial blood in superficial vessels creates characteristic thermal imprints, which are at a gradient with the surrounding tissue. It has been found that there exists a similarity between fingerprints of human beings and thermal imprints of human faces. The thermal imprints of the blood vessels may be treated as the ridges in the fingerprints and fingerprints recognition techniques [9] may be applied on thermal imprints of the human faces for their recognition. Most common type of minutiae is: when a ridge either comes to an end, which is called a ridge-termination or when it splits into two ridges, which is called a ridge-bifurcation. Fig. 1a) illustrates an example of a ridge termination and 1b) depicts an example of a ridge bifurcation point.

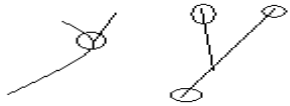

Fig. 1 a) Ridge bifurcation  b) Ridge termination.

So, the number and locations of the minutiae vary from face to face in any particular person. When a set of face images is obtained from an individual person, the number of minutiae is recorded for each face. The precise locations of the minutiae are also recorded, in the form of numerical coordinates, for each face and the resultant function thus obtained has been stored in a separate database. A computer can rapidly compare this function with that of anyone else in the world whose face image has been scanned. The paper is organized as follows: Section 2 presents about the outline proposed system. Section 3 shows the experiment and results and finally, Section 4 concludes and remarks about some of the aspects analyzed in this paper.

## 2. Outline of the proposed system

Thermal Face Recognition System (TFRS) can be subdivided into two main parts. First part is image processing and the second part is classification. The image processing part consists of thermal face image, image binarization, extracted the largest component as a face region, find the centroid of the face region, crop the face region in elliptic shape, extract the blood perfusion data and finally extract minutiae points as feature vector. These feature vectors are fed into artificial intelligent system. The second part consists of artificial intelligent system which is based on back propagation feed forward neural network. The outline of the proposed system is given in Fig. 2. The system starts with acquisition of thermal face image and end with successful classification. This successful notation comes through the application of a set of image processing and classification technique which has been discussed details in subsequent subsections.

*2.1. Thermal face image acquisition*

To collect 24-bits colour thermal face images, A FLIR 7 thermal infra red camera has been used, the images are saved in JPEG format. A typical thermal face image is shown in Fig. 3 has been given as a sample. This thermal face image depicts interesting thermal information of a facial model. Physiological features of a thermal face image have already been discussed in details in the introduction section.

*2.2. Binarization*

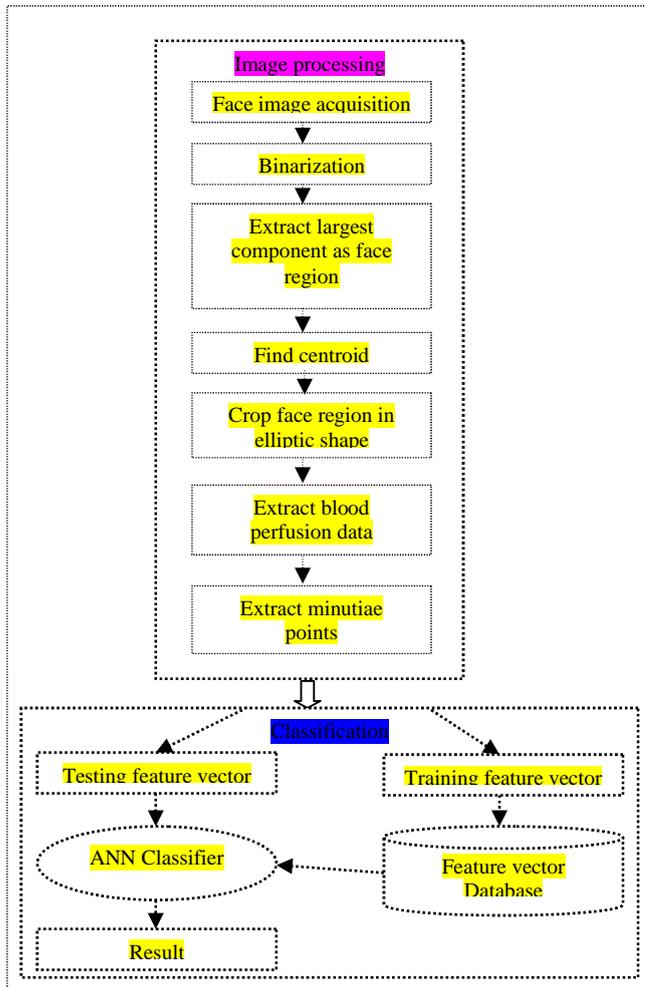

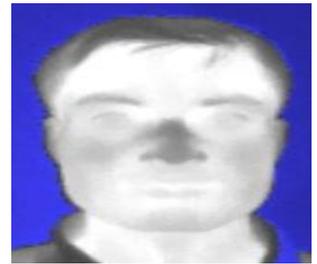

Fig. 3 A thermal infrared image

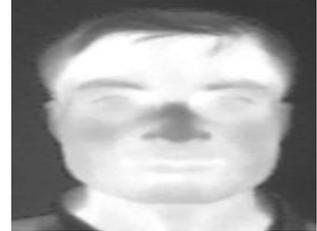

Fig. 4a) Grayscale image

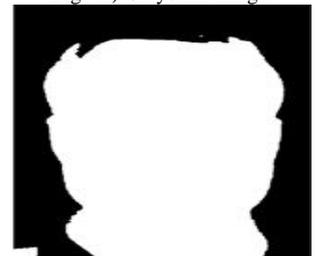

b) Binary image

Fig. 2 Schematic block diagram of the proposed system

Each of the captured 24-bits colour images have been converted into its 8-bit grayscale image. The grayscale image from the previous sample image is shown in Fig. 4a). Then convert binary image from converted grayscale image. The resultant image replaces all pixels in the grayscale image with luminance greater than mean intensity with the value 1 (white) and replaces all other pixels with the value 0 (black). In binary image, black pixels mean background and white pixels mean the face region. The binary image of the Fig. 4a) is shown in Fig. 4b).

*2.3. Extracted largest component*

A binary image may contain more than one object. Say, in Fig. 5b), it has 3 components. Then largest component has been extracted from binary image using "Connected Component Labeling" algorithm [10]. It is based on "4-conneted" neighbours or "8-connected" neighbours method [11]. A pixel to be connected to itself is called reflexive. A pixel and its neighbour are mutually connected is called symmetric. 4-connectivity and 8-connectivity are also transitive: if A is connected to pixel B, and pixel B is connected to pixel C, then there exists a connected path between pixels A and C. A relation (such as connectivity) is called an equivalence relation if it reflexive, symmetric and transitive. All equivalence classes of connected pixels in a binary image, is called connected component labelling

"Connected component labeling" algorithm is given below.
**LabelConnectedComponent(im)**
// LabelConnectedComponent(im) is a method which takes //one argument that is an image named im. f(x,y) is the //current pixel with $x^{th}$ row and $y^{th}$ column.
1. Pass the control through the whole image pixel by pixel across each row in order to get connected component.
    1.1 **if** ( f(x,y) has no associated neighbors with the same value that have already been labeled) **then**
        Create a new unique label and assign it to that
        pixel.
    1.2 **if** ( f(x,y) has exactly one label among its associated neighbor with the same value that have already been labeled) **then**
        Give it that label.
    1.3 **if** ( f(x,y) has two or more associated neighbors with the same value but different labels) **then**
        Choose one of the labels and memorize that        these labels are equivalent.
2. Go another pass through the image to determine the equivalencies and labeling each pixel with a unique label for its equivalence class.

As an illustrative example, consider a largest component as a face skin region illustrated in Fig. 8 using "Connected component labeling" algorithm, which is discussed above. It is a binary image. Here white means face skin region i.e. "1" and black means background i.e. "0".

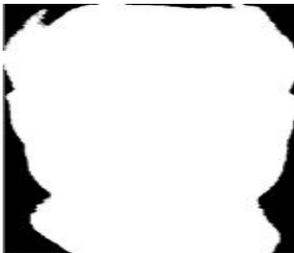

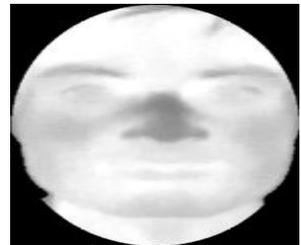

Fig. 5 A largest component as a face skin region .                                    6 Face region in elliptic shape

### 2.4. Find centroid [12]

Centroid has been extracted from the binary image in Fig. 5 using equation 1 and 2.

$$X = \frac{\sum m_{f(x,y)} x}{\sum m_{f(x,y)}} \ldots\ldots\ldots\ldots\ldots (1) \qquad Y = \frac{\sum m_{f(x,y)} y}{\sum m_{f(x,y)}} \ldots\ldots\ldots\ldots\ldots (2)$$

Where x, y is the co-ordinate of the binary image and m is the intensity value that is $m_{f(x, y)} = f(x, y) = 0$ or 1.

### 2.5. Crop the face region in elliptic shape

Normally human face is like an ellipse shape. Then from the centroid human face has been cropped in ellipse shape using "Bresenham ellipse drawing" [13] algorithm where, X and Y is x co-ordinate and y-co-ordinate respectively of the centroid which is calculated by equation 1 and 2. Distance between the centroid and the right ear is called the minor axis of the ellipse and distance between the centroid and the fore head is called major axis of the ellipse. After find the centroid of the face, face has been cropped in elliptic shape and map to the grayscale image, which is shown in Fig. 6.

### 2.6. Extract Blood perfusion data from the elliptic shaped face

Extract the blood perfusion data refers to the process of identifying the discontinuities in an image. The discontinuities are abrupt changes in pixel intensity. Here "Sobel" edge detector operator [11] is used to convolve the image for identifying discontinuities in pixel intensity. The operator consists of a pair of 3×3 convolution masks as shown in Fig. 7. One mask is simply the other flipped by 90º.

| -1 | 0 | +1 |
|---|---|---|
| -2 | 0 | +2 |
| -1 | 0  | +1 |

Px

|    |    |    |
|----|----|----|
| +1 | +2 | +1 |
| 0  | 0  | 0  |
| -1 | -2 | -1 |

Py

Fig. 7 "Sobel" edge detector masks

These masks are designed to respond maximally discontinues of intensity values running vertically and horizontally relative to the pixel grid, one mask for each of the two perpendicular orientations. The masks can be applied separately to the input image, to produce separate measurements of the gradient component in each orientation (call these P$x$ and P$y$). These can then be combined together to find the absolute magnitude of the gradient at each point and the orientation of that gradient. The gradient magnitude is given by equation 3 or 4:

$$|P| = \sqrt{Px^2 + Py^2} \ldots (3) \qquad |P| = |Px| + |Py| \ldots (4) \qquad \theta = \arctan\left(\frac{Py}{Px}\right) \ldots (5)$$

The orientation of the angle of the gradient is given by equation 5. "Sobel" edge detector mask is applied to extract lines similar to isothermal lines in whether maps linking all points of equal or constant temperature in order to get blood perfusion image. Fig. 8 shows blood perfusion data for the image shown in Fig 6.

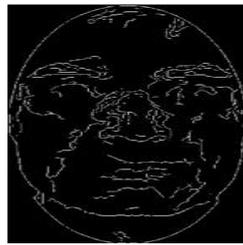

Fig. 8 Blood perfusion image

## 2.7. Extract minutiae points

The idea of minutiae extraction from finger print recognition have been taken from [9],[13],[14],[15] and applied it in the present work. There exists an analogy between thermal imprints of human faces and fingerprints of human beings. Here fingerprint's ridges are like blood perfusion data of a face. The uniqueness of a face's blood perfusion data can be determined by the pattern of ridges as well as the minutiae points. Minutiae points are local ridge characteristics that occur either at a ridge bifurcation or at a ridge termination.

| 0 | 0 | 1 |
|---|---|---|
| 0 | 1 | 0 |
| 0 | 0 | 0 |

| 1 | 1 | 0 |
|---|---|---|
| 1 | 1 | 0 |
| 0 | 0 | 0 |

| 0 | 1 | 0 |
|---|---|---|
| 0 | 1 | 1 |
| 0 | 0 | 0 |

Fig. 9a, 9b and 9c Binary number indicating the minutiae point

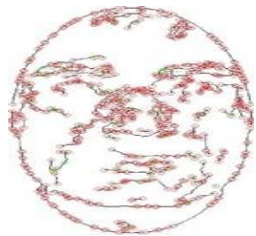

Fig. 10 Minutiae points

The number of '1''s within each 3x3 window on the blood perfusion image where minutiae points are basically the terminations and bifurcations of the ridge lines that constitutes a faceprint is computed. This is the vital part of the minutiae extraction of the faceprint image where the termination point and bifurcation point will be determined. If the central cell has a '1' and has another '1' as it's only neighbors, then it is a termination point like in Fig. 9a. If the central cell contains a '1' and has three '1''s as neighbors, then it is a bifurcation as shown in Fig. 9b and if the central cell is '1' and has two '1''s as neighbors, then it is a normal point like in Fig. 9c. After minutiae extraction the whole image is divided into number of fixed sized blocks. The size of the each block may be 8×8, 16×16, and 32×32. The number of minutiae points in each block have been counted and stored in a vector. Fig. 10 illustrates minutiae points, which are basically the bifurcation and termination points represented by green and red colors respectively. For each face image, one corresponding vector have been found i.e. total number of such vectors are equal to total number of faces. Then divide these vectors into two sets one for training purpose and another for testing purpose. ANN classifier has been used to classify each of these vectors [16],[17],[18]. Here a five layer feed-forward back propagation neural network has been used for classification purpose. First hidden layer contains 100 neurons, second hidden layer contains 50 neurons, third hidden layer holds 10 neurons, input layer contains 238 neurons for training of 238 images and the last layer contains 7 neurons because 7 different persons are there in our experiment. Tan-sigmoid transfer function has been used as gradient descent with momentum training function is used to update weights and bias values in this network.

## 3. Experiment and Results

The proposed algorithm is tested on thermal face images captured using a FLIR 7 camera at our own laboratory. Some preprocessing has been done for the image database used in this paper. All the training and testing images are in color with 24-bits. Face images of a person was captured in normal temperature conditions, and each person sitting on a chair at a distance of about 2 feet in front of the thermal IR camera. Each person has 34 different templates: emotion type without changing head orientation, different views about y axis, different views about x axis, and different views about z axis. The obtained results are shown in Table 1 for different block sizes.

Table 1. Performance rate for different block size

| No of Block | Performance Rate |
|---|---|
| 8×8 | 90.25 |
| 16×16 | 88.10 |
| 32×32 | 83.33 |

Over the last few years, many researchers have investigated the use of thermal infrared face images for person identification (Chen et al. 2003; Buddharaju et al., 2004; Socolinsky & Selinger, 2004, Singh et al., 2004, Buddharaju et al., 2007). The techniques developed by Socolinsky & Selinger (Socolinsky & Selinger, 2004) show performance statistics for outdoor face recognition and recognition across multiple sessions. A few experimental results with thermal images in face recognition are being recorded in Table 2. All the result support the conclusion that face recognition performance with thermal infrared imagery is stable over multiple sessions.

Table 2. Comparison of recognition rate of different methods.

| Method | Recognition rate |
|---|---|
| Fusion of Thermal and Visual (Singh et al. 2004) | 90% |
| Segmented Infrared Images via Bessel forms (Buddharaju et al., 2004) | 90% |
| PCA for Visual indoor Probes (Socolinsky & Selinger, 2004) | 81.54% |
| PCA+LWIR (Indoor probs) (Socolinsky & Selinger, 2004) | 58.89% (Maximum) |
| LDA+LWIR (Indoor probs) (Socolinsky & Selinger, 2004) | 73.92% (Maximum) |
| PCA+LWIR (Outdoor probs) (Socolinsky & Selinger, 2004) | 44.29% (Maximum) |
| LDA+LWIR (Outdoor probs) (Socolinsky & Selinger, 2004) | 65.30% (Maximum) |
| Equinox+LWIR (Outdoor probs) (Socolinsky & Selinger, 2004) | 83.02% (Maximum) |

## 4. Conclusions

Minutiae based thermal face recognition using blood perfusion data has been proposed here. Feature extraction is done by using some feature extraction technique, which is discussed earlier. Classification of these feature vectors has been done using a multilayer perceptron. Final recognition rate has been enhanced by varying size of the blocks. One of the major advantages of this approach is the ease of implementation. Future work includes tests with more images with the proposed approach.

## Acknowledgements

Authors are thankful to a major project entitled "Design and Development of Facial Thermogram Technology for Biometric Security System," funded by University Grants Commission (UGC),India and "DST-PURSE Programme" at Department of Computer Science and Engineering, Jadavpur University, India for providing necessary infrastructure to conduct experiments relating to this work.

## References

1. D. A. Socolinsky and A. Selinger, "A Comparative Analysis of Face Recognition Performance with Visible and Thermal Infrared Imagery," Proc. Int. Conf. on Pattern Recognition, Vol. 4, pp.217-222, Quebec, 2002.
2. Socolinsky, Selinger, "Face recognition with visible and thermal infrared imagery", 2003.
3. Shiqian Wu, Zhi-Jun Fang, Zhi-Hua Xie and Wei Liang, "Blood Perfusion Models for Infrared Face Recognition" School of information technology, Jiangxi University of Finance and Economics, China.
4. S.G. Kong, J. Heo, B.R. Abidi, J. Paik, M.A. Abidi, Recent advances in visual and infrared face recognition—a review, Comput. Vision Image Understanding 97 (2005) 103–135.
5. Buddharaju, P.; Pavlidis I. & Kakadiaris I. A. (2004). Face recognition in the thermal infrared spectrum, Proceedings of IEEE International Conference on Computer Vision and Pattern Recognition Workshop, pp. 133-133, Washington DC, USA, 2004.
6. C. Manohar, "Extraction of Superficial Vasculature in Thermal Imaging," master's thesis, Dept. Electrical Eng., Univ. of Houston, Houston, Texas, Dec. 2004.
7. Guyton, A. C. and Hall, J. E. (1996). Textbook of Medical Physiology, 9th ed., Philadelphia: W.B. Saunders Company, 1996.
8. F. Prokoski, History, current status, and future of infrared identification, in: Proceedings of the IEEE Workshop Computer Vision Beyond Visible Spectrum: Methods and Applications, 2000, pp. 5–14.
9. Anil Jain, Arun Ross, and Salil Prabhakar, Fingerprint matching using minutiae and texture features,in Proc. of Intnl. Conf. on Image Processing ICIP,pp. 282-285, Thessaloniki, Greece, Oct. 7-10, 2001.
10. Bryan S. Morse, Lecture 2: "Image Processing Review, Neighbors, Connected Components, and Distance", 1998-2004.
11. R.C. Gonzalez and R.E. Woods, Digital Image Processing, 3rd Edition, Prentice Hall, 2002.
12. Venkatesan S.and Madane S.S.R.," Face Recognition System with Genetic Algorithm and ANT Colony Optimization", International Journal of Innovation, Management and Technology, Vol. 1, No. 5, December 2010.
13. D. Hearn M. P.Baker Computer graphics C version.
14. F. Galton. Finger Prints. Mcmillan, London, 1892.
15. Davide Maltoni, Dario Maio, Anil K. Jain, Salil Prabhakar; Handbook of Fingerprint Recognition, second Edition, Springer-Verlag London Limited, 2009.
16. M. Turk and A. Pentland, "Eigenfaces for recognition", Journal of Cognitive Neuroscience, vol.3, No.1, 1991.
17. M. Turk and A. Pentland, "Face recognition using eigenfaces", *Proc. IEEE Conf. on Computer Vision and Pattern Recognition*, pp. 586-591, 1991.
18. Lin and Lee: Neural Fuzzy Systems: Prentice Hall International ,1996.